\documentclass{article}
\usepackage{ijcai24}

\usepackage{times}
\usepackage{soul}
\usepackage{url}
\usepackage[hidelinks]{hyperref}
\usepackage[utf8]{inputenc}
\usepackage[small]{caption}
\usepackage{graphicx}
\usepackage{amsmath}
\usepackage{amsthm}
\usepackage{booktabs}
\usepackage{algorithm}
\usepackage{algorithmic}
\usepackage[switch]{lineno}

\usepackage{color}
\usepackage{multirow}
\usepackage{amssymb}

\newcommand{\methodname}{{\tt{IA-AFL}}}

\title{Intelligent Agents for Auction-based Federated Learning: A Survey}

\author{
Xiaoli Tang$^1$
\and
Han Yu$^1$
\and
Xiaoxiao Li$^2$
\And
Sarit Kraus$^{1,3}$
\affiliations
$^1$School of Computer Science and Engineering, Nanyang Technological University, Singapore\\
$^2$Department of Electrical and Computer Engineering, The University of British Columbia, Canada\\
$^3$Department of Computer Science, Bar-Ilan University, Israel\\
\emails
\{xiaoli001, han.yu\}@ntu.edu.sg, xiaoxiao.li@ece.ubc.ca, sarit@cs.biu.ac.il
}

\begin{document}

\maketitle

\begin{abstract}
    Auction-based federated learning (AFL) is an important emerging category of FL incentive mechanism design, due to its ability to fairly and efficiently motivate high-quality data owners to join data consumers' (i.e., servers') FL training tasks. To enhance the efficiency in AFL decision support for stakeholders (i.e., data consumers, data owners, and the auctioneer), intelligent agent-based techniques have emerged. However, due to the highly interdisciplinary nature of this field and the lack of a comprehensive survey providing an accessible perspective, it is a challenge for researchers to enter and contribute to this field. This paper bridges this important gap by providing a first-of-its-kind survey on the \underline{I}ntelligent \underline{A}gents for \underline{AFL} (\methodname{}) literature. We propose a unique multi-tiered taxonomy that organises existing \methodname{} works according to 1) the stakeholders served, 2) the auction mechanism adopted, and 3) the goals of the agents, to provide readers with a multi-perspective view into this field. In addition, we analyse the limitations of existing approaches, summarise the commonly adopted performance evaluation metrics, and discuss promising future directions leading towards effective and efficient stakeholder-oriented decision support in \methodname{} ecosystems.
    
\end{abstract}

\section{Introduction}
\label{sec:introduction}
Federated Learning (FL) is a collaborative machine learning (ML) paradigm that is able to train useful models while respecting user privacy and data confidentiality \cite{yang2019federated,konevcny2016federated,zhang2021incentive}. FL has gained significant attention from academia \cite{yang2019federated} and industry \cite{Liu-et-al:2020FedVision,Liu-et-al:2022IAAI} alike, leading to a diverse range of techniques \cite{Kairouz-et-al:2021}. In FL, there are two types of participants: \textit{data consumers} (DCs, who often perform the role of FL servers), overseeing the distribution and aggregation of global FL models, and \textit{data owners} (DOs, who often play the role of FL clients), responsible for training the FL model using their local data. FL follows a distributed approach where each DO trains a local model on its private dataset, and shares it with the corresponding DC. The DC then aggregates the received local models following an aggregation algorithm (e.g., FedAvg \cite{mcmahan2017communication}) to obtain the global model, which is then distributed back to the DOs for further training until convergence criteria are met. This design ensures that private local data are not exposed to any party other than the original owner, thus reducing privacy risk.

Despite these advantages, existing FL works generally assume that all DOs agree to participate in the FL training process when requested \cite{le2021incentive}. However, in practice, DOs are self-interested entities who consider a complex set of factors (e.g., costs, potential risks of privacy exposure, expected utility gains) before deciding to join an FL task. This has motivated the study of FL incentive mechanisms \cite{khan2020federated}, which aims to develop effective mechanisms that align the interests of DOs with the goals of DCs. They play a crucial role in encouraging DOs to actively participate in FL and make valuable contributions, ultimately leading to improved performance and broader adoption of FL in real-world applications.

\begin{figure}[t!]
\centering
\includegraphics[width=0.95\columnwidth]{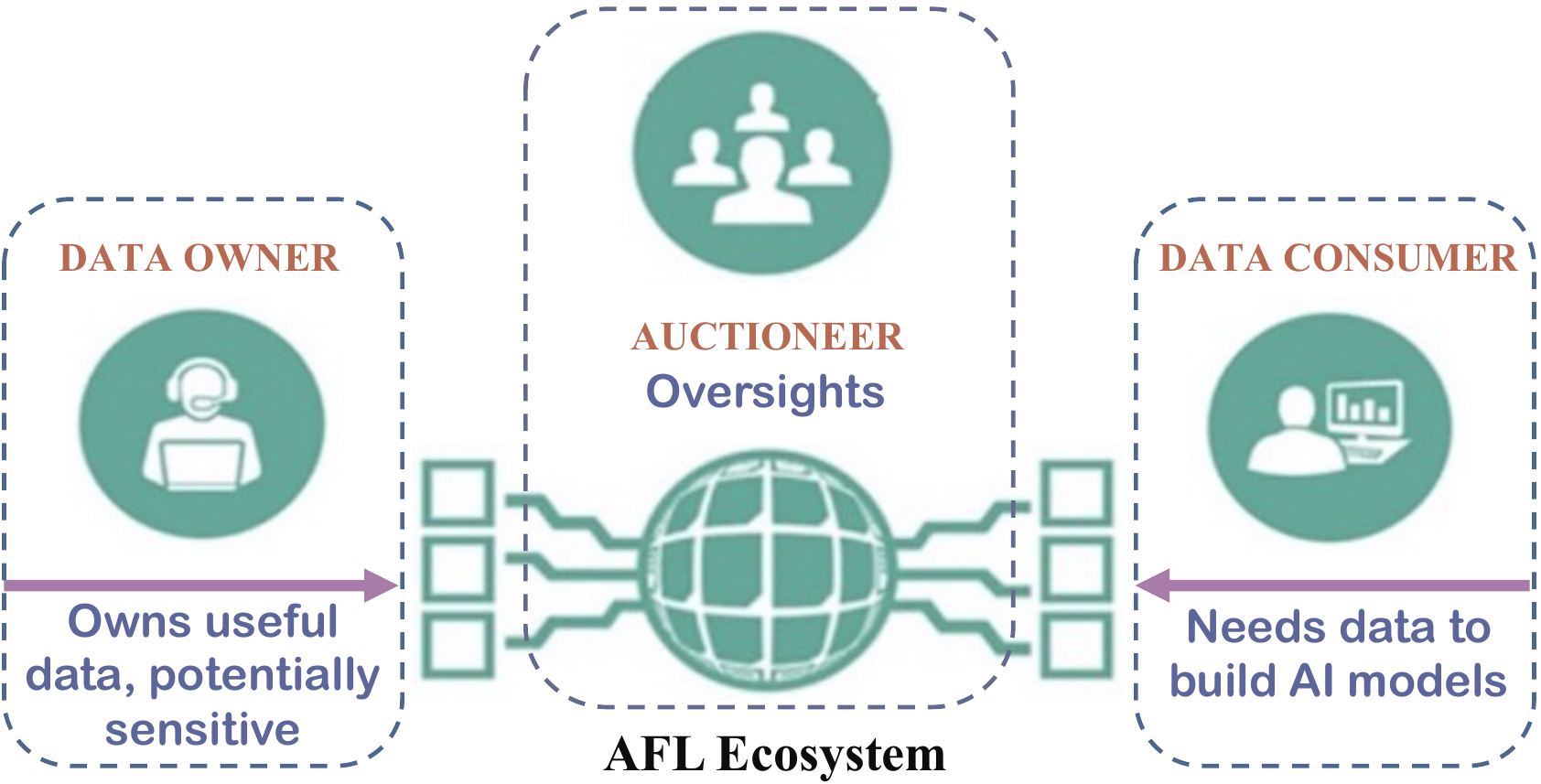}
\caption{An overview of the AFL ecosystem.}
\label{fig:3_side_problem}
\end{figure}

Auction-based approaches have gained significant attention recently as an effective way to design FL incentive mechanisms. They offer a promising approach to motivating DOs to participate in FL in a fair and efficient manner. Under the typical auction-based FL (AFL) setting\footnote{A possible example open AFL marketplace can be the Hierarchical Auctioning in Crowd-based Federated Learning system \cite{Gao-et-al:2023ICME}: https://hacfl.federated-learning.org/.}, three key stakeholders are involved: 1) DCs, 2) DOs, and 3) an auctioneer (as illustrated in Figure \ref{fig:3_side_problem}). The auctioneer plays a crucial role in coordinating the auction process, while DOs and DCs provide the auctioneer with their available data resources and bid values, respectively. The auction process as well as the entire AFL ecosystem center around the decision-making process of each stakeholder. The decisions made by each stakeholder impact the outcomes of AFL. To deal with the complexity, dynamism and personal nature of the context and the decision-making process, intelligent agents are often adopted to provide these stakeholders with AFL decision support, thereby inspiring the field of \underline{I}ntelligent \underline{A}gents for \underline{AFL} (\methodname{}).

\methodname{} is highly interdisciplinary in nature. It requires expertise from machine learning, multi-agent systems, game theory and auction theory, etc. This makes it challenging for researchers new to the field to grasp the latest developments. Currently, there is no survey paper on this important and rapidly developing field. To bridge this gap, we conduct a comprehensive survey of research works focusing on \methodname{} in this paper.\footnote{Although some of the papers included in this survey do not explicitly mention agents, their focus on providing decision support for stakeholders in AFL reflects their potential as useful building blocks for realizing an agent-based AFL system.} We analyse the AFL ecosystem in detail, with a focus on the diverse stakeholders involved and their decision-making priorities. Based on this analysis, we propose a unique multi-tiered taxonomy of \methodname{} that organises existing works according to 1) the stakeholders served, 2) the auction mechanism adopted, and 3) the goals of the agents to provide readers with a multi-perspective view into this field. In addition, we analyse the limitations of existing approaches, summarise the commonly adopted performance evaluation metrics, and discuss promising future directions towards effective and efficient stakeholder-oriented decision support in \methodname{} ecosystems.

\section{Preliminaries}
\label{sec:pre}
\subsection{A Typical AFL Ecosystem}
As shown in Fig. \ref{fig:3_side_problem}, a typical AFL ecosystem involves three primary stakeholders \cite{tang2023utility}: 1) DOs, who act as the sellers possessing potentially sensitive but valuable data and training resources; 2) DCs, who act as buyers of such data and training resources to build ML models; and 3) an auctioneer, overseeing the matching of DOs with DCs and providing essential governance oversight for the ecosystem.

DCs submit their bidding profiles (including the bidding prices and their FL tasks) to the auctioneer. DOs submit their asking profiles (including the FL tasks they are able to join and their asking prices) to the auctioneer. The auctioneer determines the winners, and the corresponding market prices based on the submitted asking profiles and the bidding profiles under a predefined auction mechanism, and informs the winners. The winning DCs then pay the DOs. Through such an auction process, each DC recruits DOs to join its FL task. Afterward, each DC orchestrates the FL model training process with its recruited DOs following an adopted FL protocol.

\subsection{AFL Stakeholder Concerns}
In the AFL ecosystem, the stakeholders play distinct roles with different interests and concerns.

The \textbf{auctioneer}'s role is pivotal, overseeing the auction process and facilitating information flow between participating DOs and DCs. Its main focus is to maintain the sustainable operation of the AFL ecosystem by attracting and retaining more participants, optimizing key performance indicators for the entire ecosystem, and providing governance oversight.

\textbf{Data consumers}, acting as buyers in the auction market, are primarily concerned with effective selection or bidding for DOs to meet their key performance indicators (KPIs), while staying within budget constraints.

\textbf{Data owners}, acting as sellers in the auction market, prioritize maximizing their monetary rewards. They are also keen on safeguarding data privacy by optimizing data resource allocation and the setting of reserve prices (i.e., the minimum acceptable price for selling the corresponding data resources). 

\subsection{Terminology}
For ease of understanding, we provide a brief overview of key terminology adopted by the AFL field:

    \textbf{Commodity / data resources}: In AFL, the term commodity refers to the object being exchanged between DCs and DOs, denoting a specific value for buying or selling. It can represent a unit of data (e.g., a training sample), communication bandwidth committed by a DO, or a unit of compute resource. In this paper, we use the terms data resources and commodity interchangeably unless a specific distinction is necessary.
    
    \textbf{Valuation}: Valuation in AFL involves the assessment of the monetary value of data resources. Different DCs and DOs may assign value to data resources differently based on their individual preferences. Valuation can be either private, undisclosed to others, or public.
    
    \textbf{Utility}: For DCs, utility is defined as the difference between their valuation of the auctioned data resources and the eventual payment made for those resources. For DOs, utility is defined as the difference between the total payments received from DCs and the costs incurred for the data resources, including communication and computation costs.
    
    \textbf{Social welfare (SW)}: SW is the sum of utilities for some or all participants in an AFL ecosystem. It provides a measure of the collective benefit derived from all transactions.

\subsection{Types of Auction}
AFL ecosystems can adopt various auction mechanisms based on their specific application scenarios \cite{qiu2022applications}, including 1) double auction, 2) combinatorial auction, 3) reverse auction, and 4) forward auction.
Double auctions \cite{friedman2018double} accommodate multiple DOs and DCs, with both sides submitting asks and bids to the auctioneer. Combinatorial auctions \cite{de2003combinatorial} are effective when DCs bid for data resource bundles, ideal for acquiring complementary data types. Reverse auctions \cite{parsons2011auctions} involve DOs competing for FL tasks, while forward auctions involve DCs competing for data resources.

Winner determination and pricing methods in AFL auctions fall into three categories \cite{tu2022incentive}: 1) 
first-price sealed-bid (FPSB), 2) second-price sealed-bid (SPSB), and 
3) Vickrey Clarke-Groves (VCG).
Under FPSB, the highest bidder wins the auction and pays the bid price. The simplicity of FPSB might lead to inefficiencies and overpayment. Under SPSB, the highest bidder wins the auction, but pays the second-highest bid price. SPSB encourages truthful bidding to reveal true item valuation.
Under VCG, winners are determined by maximizing the total benefit, considering externalities. Payments are determined based on the value contributed by other bidders for efficient and accurate price discovery \cite{vickrey1961counterspeculation}.

\section{The Proposed \methodname{} Taxonomy}
\begin{figure}[b!]
\centering
\includegraphics[width=1\columnwidth]{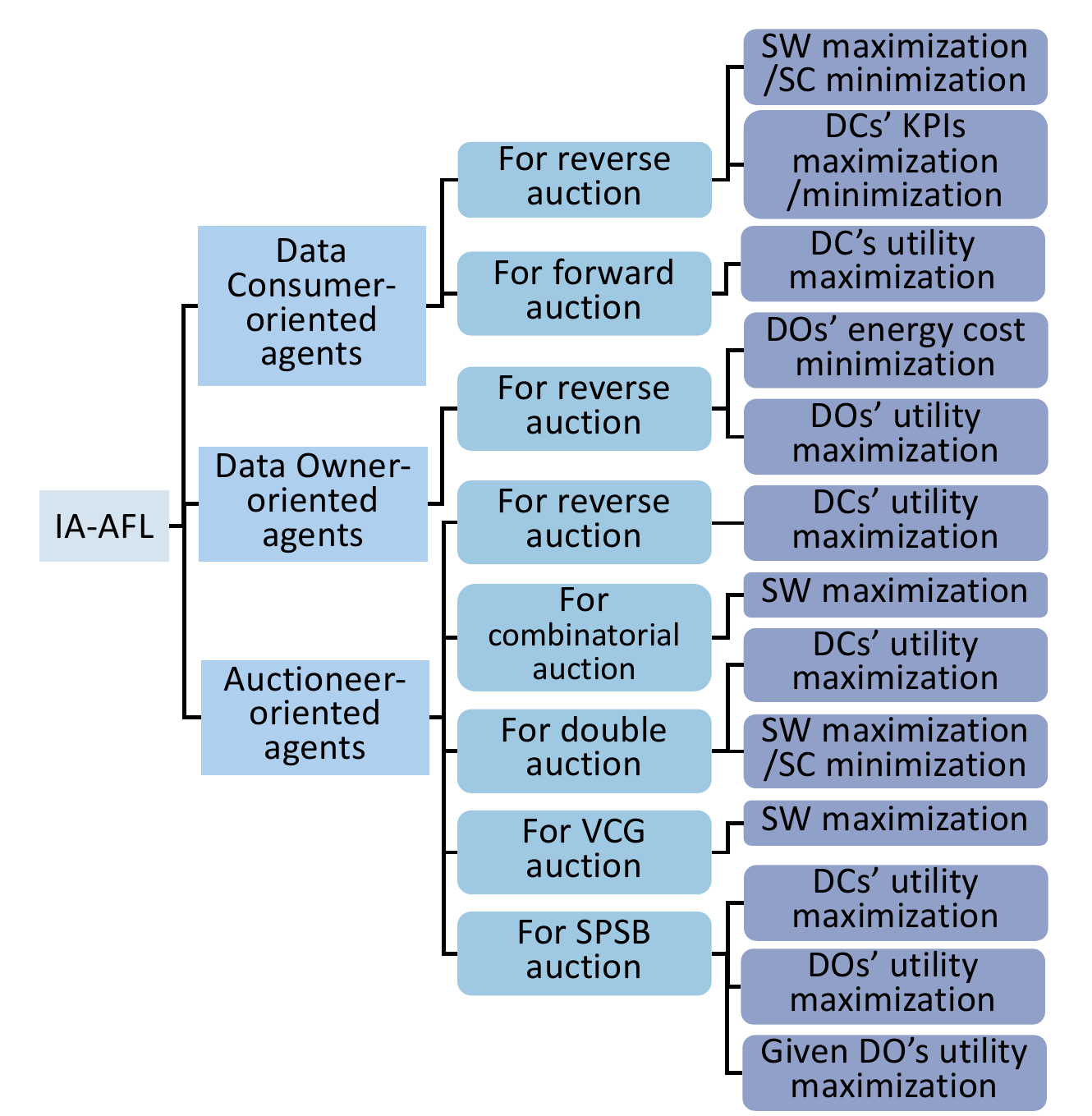}
\caption{The \methodname{} taxonomy. DC, DO, SW and SC denote data consumer, data owner, social welfare and social cost, respectively.
}
\label{fig:taxonomy}
\end{figure}

Based on the stakeholders, the types of auctions involved in AFL and their respective goals, we propose a taxonomy for the \methodname{} literature as shown in Figure \ref{fig:taxonomy}. Specifically, it first separates \methodname{} literature into data consumer-oriented, data owner-oriented, and auctioneer-oriented methods. Since all auction mechanisms introduced in the last section can be adopted by the AFL process, we further classify \methodname{} works based on their respective adopted auction mechanisms. Then, as stakeholders can have different goals, we further divide \methodname{} works based on their objectives. This hierarchical taxonomy provides a clear overview of the current \methodname{} landscape.

\subsection{Intelligent Agents for Data Consumers}
Based on the adopted auction mechanism, DC-oriented \methodname{} works can be broadly categorized into two distinct groups: 1) those designed for reverse auctions, and 2) those designed for forward auctions. These agents are instrumental in facilitating strategic decision-making for DCs, ensuring effective participation in the AFL market while maximizing key performance indicators (KPIs) derived from the collaborative FL model training process.

\subsubsection{For Reverse Auction}
Under reverse auction, existing methods assume that there is only one DC and multiple DOs in the AFL marketplace.
The intelligent agent for the DC plays a crucial role in selecting DOs. It makes decisions by evaluating DOs' asking profiles, assessing their potential contributions to the model, and aligning with the DC's objectives.
Existing \methodname{} works for DCs under reverse auction can be broadly classified into two categories based on their designed objectives: 1) social welfare / social cost optimization approaches, and 2) DC KPI optimization approaches.

\textbf{Social welfare / social cost optimization}: 
To optimize the social welfare objective, \cite{jiao2020toward} first groups DOs based on Earth Mover's Distances (EMD) \cite{zhao2018federated}. The DC then greedily selects DOs from each group, determining payments based on marginal virtual social welfare density. To enhance social welfare, the authors incorporate a graph neural network to manage relationships among DOs, and use deep reinforcement learning to determine the winning DOs and their payments.
In \cite{le2020auction}, the workflow is similar, with a key distinction in the formulation of the DO selection process as a social cost minimization problem.

However, these works primarily focus on DO selection and payment determination over a single FL communication round. In \cite{zhou2021truthful}, the DC is assisted in selecting and paying DOs for different FL communication rounds. The work decomposes the social cost minimization problem into a series of winner determination problems (WDPs) based on the number of global FL iterations. Each WDP is solved using a greedy algorithm to determine winning DOs, and a payment algorithm for computing remuneration to the winners. In \cite{yuan2021incentivizing}, the focus is on social cost minimization over the long run. The proposed FLORA method utilizes multiple polynomial-time online algorithms, including a fractional online algorithm and a randomized rounding algorithm, to select winning DOs and control the training accuracy of the global FL model. It also includes a payment algorithm to assist the DC in decision-making regarding DO selection and payment determination.

Different from the above two methods, which are designed for social cost minimization, \cite{wu2023long} focuses on social welfare maximization. To achieve this goal, the proposed method follows deep reinforcement learning to select DOs and determine their payments under the VCG auction.

\textbf{Data consumer KPI optimization}: In \cite{fan2020hybrid}, the proposed method DQDRA maximizes the DC's valuation by determining winning DOs and the corresponding payments with a monotone greedy algorithm after receiving asking profiles from all DOs. 
Unlike DQDRA, which requires knowledge about the global distribution of all data for winning DO determination, RRAFL proposed in \cite{zhang2021incentive} leverages blockchain and reputation mechanisms instead. Winning DOs are selected based on their respective reputation, which are evaluated through historical contributions to the global FL model stored on the blockchain.
Expanding on this, \cite{zhang2022auction} enhances RRAFL by introducing a novel contribution evaluation method using weighted samples. This adds nuance to the evaluation process, potentially offering a more sophisticated understanding of individual DOs' contributions. In \cite{zhang2022online}, RRAFL is extended by segmenting FL training tasks into multiple time steps based on global iterations, allowing adaptation to online learning applications.
In \cite{zeng2020fmore}, the proposed method FMORE helps the DC select the top $K$ DOs with the highest score using the Lagrange multiplier method. \cite{batool2022fl,batool2023block} follow a similar method by incorporating blockchains \cite{kang2020reliable,kim2019blockchained} and contract theory \cite{kang2019incentive} to select DOs.

The aforementioned reputation-based DO selection methods do not explicitly consider the quality of the DOs, which is crucial for FL model performance. To address this limitation, \cite{deng2021fair} proposed FAIR, which integrates a quality-aware model aggregation algorithm with the reverse auction mechanism. FAIR determines winning DOs using a greedy algorithm based on Myerson's theorem \cite{myerson1998population} to maximize the DC's valuation.

Unlike methods determining winning DOs and the corresponding payments in one communication round with a given budget, \cite{yang2023bara,tan2023reputation,tan2023hire} study how to allocate the DC's budget across multiple global FL communication rounds. \cite{yang2023bara} proposed BARA, an online reward budget allocation algorithm based on Bayesian optimization. Considering the urgency of recruitment, \cite{tan2023reputation,tan2023hire} help the DC determine time-averaged optimal budget allocation for DOs.

\textbf{Limitations}: Existing works in this area often operate under the assumption of a monopolistic AFL market, where multiple DOs vie to join the FL training tasks of a single DC. However, this assumption diverges from the reality of practical AFL marketplaces, where numerous DCs may compete to attract multiple DOs for their respective FL training tasks.

\subsubsection{For Forward Auction}
Works in this field focus on maximizing the utility of a given DC within an AFL marketplace, which often involves multiple DCs.
In \cite{tang2023utility}, a utility-maximizing bidding strategy, FedBidder, is designed for the DCs. It leverages various auction-related insights (e.g., DOs' data distributions, suitability to the task, DCs' bidding success probabilities, and budget constraints). The study emphasizes the crucial roles played by the estimation of DOs' utility and the appropriate winning function design in determining the optimal bidding function. To solve the optimal bidding function effectively, a utility estimation algorithm was proposed with two representative winning functions introduced, deriving two forms of optimal bidding functions for the DCs.

However, this approach overlooks the intricate relationships among DCs, which can be simultaneously competitive and cooperative. To address this issue, researchers have explored incorporating more than one agent for each DC.
In \cite{tang2023competitive}, the AFL ecosystem is modeled as a multi-agent system to guide DCs in strategically bidding towards an equilibrium with desirable overall system characteristics. The proposed approach, MARL-AFL, assigns two agents to each DC: 1) a bidding agent for determining bid prices, and 2) a bar agent for setting the bidding lower bound for the corresponding bidding agent. The bar agent is introduced to address potential collusive behaviors among bidding agents, such as bidding with an extremely low bid price, which can be detrimental to the health of the entire ecosystem. Both the bidding agents and the bar agents are designed based on deep Q-networks (DQN) \cite{mnih2015human}.

In \cite{tang2023multi}, MultiBOS-AFL is proposed to assist the DC in bidding for DOs in competitive AFL marketplaces. Unlike FedBidder and MARL-AFL, which assume that the entire team of DOs required for an FL task must be assembled before training can commence, MultiBOS-AFL helps the DC bid for DOs gradually over multiple FL model training sessions. To achieve this goal, each DC is assigned two agents: one for optimizing inter-session budget pacing, and the other for optimizing intra-session bidding. 

\textbf{Limitations}: In this area, existing studies often assume that DOs arrive sequentially before the auction begins. However, real-world scenarios frequently involve DOs arriving in diverse orders, either before or during FL training tasks. The current body of research lacks robust solutions to navigating these dynamic and evolving situations effectively.

\subsection{Intelligent Agents for Data Owners}
In AFL, DOs function as the sellers, offering their valuable data resources to DCs. This transaction leads them to eventually become participants in the FL training processes initiated by various DCs, with the prospect of receiving monetary rewards in return. Consequently, intelligent agents tailored for DOs play a crucial role in providing guidance on strategic decision-making related to the allocation of their data resources and determining the asking profiles for these resources. Their final objective is to optimize the monetary profits derived from their involvement in AFL.

\subsubsection{For Reverse Auction}
\textbf{Data owner energy cost minimization}:
In \cite{le2021incentive}, the data resource trading process between a data consumer and multiple data owners is modeled as a reverse auction. Upon receiving FL training task profiles from the data consumer, which include the maximum tolerable time for FL training, each data owner optimizes asking profiles. These profiles, encompassing parameters like uplink transmission power, local accuracy level, and CPU cycle frequency, are fine-tuned iteratively to minimize energy costs.

\textbf{Data owner utility maximization}:
In \cite{lu2023auction}, a within-cluster DO selection scheme was proposed for reverse auction to address the problem of uneven data resource consumption in a given cluster. DOs determine bid prices by maximizing their total utility. Similarly, \cite{le2020auction} also focuses on maximizing DO utility. However, unlike \cite{lu2023auction} which solves the utility maximization problem to obtain bid prices, \cite{le2020auction} aims to derive asking profiles including CPU cycle frequency, uplink transmission power and training costs, in order to maximize utility.
In \cite{zeng2020fmore}, when a DO receives an FL training task and a scoring function from the DC, the proposed method assists it in deciding whether to bid based on its available data resources. If the DO chooses to bid, decisions regarding the number of resources to allocate and the corresponding charges to the DC are made using Euler's method.

\textbf{Limitations}: To the best of our knowledge, only these four studies currently address the issue of agent-based DO decision support. However, each of these works only concentrates on a single aspect of a DO's concerns. In practice, each decision made by a DO should encompass multiple facets simultaneously to meet its KPIs. Focusing solely on one aspect may lead to sub-optimal solutions.

\subsection{Intelligent Agents for the Auctioneer}
In an AFL ecosystem, the auctioneer serves as the coordinator and administrator, overseeing the flow of information between DOs and DCs, and facilitating the trading processes. Therefore, intelligent agents designed for the auctioneer are pivotal in offering strategic guidance for matching DOs and DCs. The ultimate goal is to optimize the monetary profits derived from their engagement within the AFL ecosystem. Existing methods in this domain are designed for four main auction mechanisms: 1) reverse auction, 2) combinatorial auction, 3) double auction, and 4) VCG/SPSB auction.

\subsubsection{For Reverse Auction}
\textbf{Data consumer utility maximization}:
In \cite{seo2021auction}, the auctioneer, represented by the software-defined network controller, facilitates decision-making between the DC and DOs. It determines the minimum number of global communication rounds required to meet the quality requirements of the FL model. This decision-making process occurs within the context of a reverse auction-based data trading system. Similarly, in \cite{seo2022resource}, a software-defined network controller serves as the auctioneer, positioned between the DC and DOs. The proposed method in this paper assists the auctioneer in making decisions during the selection of winning DOs. The objective is to maximize the utility of the DC, via a greedy method.

\textbf{Limitations}: Like the \methodname{} approaches designed for the DC under reverse auction, these methods also operate under the assumption of a monopolistic AFL market. This assumption might constrain the practical applicability of these methods in real-world scenarios.

\subsubsection{For Combinatorial Auction}
\textbf{Social welfare maximization}: \cite{xu2023cab} aims to maximize social welfare and protect the utility of the auctioneer. The approach involves two main stages: 1) the combinatorial auction stage, where the platform selects winners who make the total utility of the platform and themselves greater than zero, and 2) the bargaining stage, where winners are classified into two categories with different payment methods after completing the training model. The goal is to ensure the utility of the auctioneer remains positive.

\textbf{Limitations}: \cite{xu2023cab} operates under the premise of a monopoly AFL market, assuming a single platform orchestrating the auction processes. While this setting provides a basis for understanding, a critical challenge lies in expanding participation, particularly attracting more DOs to engage in AFL. Enticing a diverse range of participants and optimizing the platform's functionality under more realistic, competitive scenarios remains an open area for exploration. 

\subsubsection{For Double Auction}
Under double auction settings, the auctioneer agent ultimately coordinates agents serving DOs and DCs. Therefore, they are treated as auctioneer agents by extension.

\textbf{Data consumer utility maximization}: 
FEST \cite{roy2021distributed} matches DOs and DCs with the goal of maximizing DC utility. This utility is a composite function involving the DC's valuation for data resources, the DO's asking price, and the corresponding execution time and reputation value. FEST assist DOs in determining winning candidate DCs using a greedy approach, followed by helping DCs select DOs to maximize their utility.

\textbf{Social welfare / social cost optimization}: 
\cite{mai2022automatic} assists the auctioneer in matching DCs and DOs, with the aim of maximizing social welfare. DOs submit asking profiles, and DCs submit bidding profiles to the auctioneer,

which, in turn, uses the Lagrangian function to perform DO-DC matching.
In \cite{wang2023toward}, the focus is on social cost minimization under double auction. The authors formulate a nonlinear mixed-integer program for long-term social cost minimization. They propose an algorithmic approach to generate candidate training schedules and solve the problem using an online primal-dual-based algorithm \cite{buchbinder2009design} with a carefully embedded payment design.

\textbf{Limitations}: Current methods predominantly operate under a centralized framework, where a central server continuously aggregates global system information and computes optimal decisions for the auctioneer. While the merits of a centralized architecture, such as rapid convergence and global optimality, are evident, they come at the cost of significant communication and computation overhead, especially in large-scale AFL ecosystems. Whenever there are shifts in the requirements of DCs, the auctioneer must collect extensive information across the entire ecosystem and recompute decisions. Moreover, in the event of hardware failures or attacks on the auctioneer, the entire ecosystem can be compromised.

\subsubsection{For VCG Auction}
\textbf{Social welfare maximization}: 
FVCG \cite{cong2020vcg} helps the auctioneer determine the amount of acceptable data to maximize its utility, factoring in data quality and privacy cost from DOs. It adopts a composite neural network-based payment function to derive payments for each DO, aiming to maximize social welfare and ensure fairness among DOs.
Extending FVCG, \cite{cong2020optimal} introduced PVCG, which incorporates a game-theoretical model for the co-creation of virtual goods. PVCG helps the auctioneer determine the acceptance of input resources from each DO based on its asking profile, and imposes penalties if it fails to deliver the claimed resources. The objective is to maximize social welfare and mitigate information asymmetry.

\textbf{Limitations}: As the number of DOs increases, the need for more effective and efficient models to learn how to compensate DOs effectively becomes apparent for both FVCG and PVCG. Furthermore, it is crucial to evaluate the effectiveness of FVCG and PVCG in comparison to other sharing rules, such as Shapley value \cite{liu2022gtg} and labour union \cite{gollapudi2017profit}.

\subsubsection{For SPSB Auction}
\textbf{Data consumer utility maximization}: 
In \cite{xu2021bandwidth}, a multi-bid auction mechanism is introduced to address bandwidth allocation challenges for self-interested DCs. The primary objective is to maximize the utility of DCs. Under this method, DCs submit bidding profiles specifying their requested bandwidth and unit price to the auctioneer. The auctioneer then allocates the bandwidth to DCs based on the market clearing price, and each DC incurs charges according to the SPSB auction mechanism.

\textbf{Data owner utility maximization}: 
In \cite{lim2020incentive}, the focus is on multiple DCs engaging in competitive bidding for data resources from a specific DO. The bids from DCs undergo a transformation, and the winning DCs are selected, with payments determined using the SPSB auction mechanism. The overarching objective is to maximize the utility of the DO.
\cite{ng2020communication,ng2020joint} incorporate Unmanned Aerial Vehicles (UAVs) as wireless relays to enhance communication between DOs and DCs. The optimal coalitional structure between UAV coalitions and DO coalitions is determined through the SPSB auction, aiming to maximize the utility of the UAV coalitions.

\textbf{Limitations}: Existing works in this area operate under the assumption that a DO can participate in at most one FL training task at any given time. In practice, DOs may have spare capacities to engage in multiple FL tasks concurrently. In such cases, resource allocation strategies should consider both the bandwidth and computing resources of the DOs. Exploring and adapting auction mechanisms to address the complexities arising from DOs' simultaneous involvement in multiple FL tasks is an open research question.

\section{Evaluation Methodology}
To assess \methodname{} methods, a combination of theoretical analysis and experimental evaluation is commonly adopted. 

\subsection{Theoretical Analysis}
Given the nature of the auction and the emphasis on incentive mechanisms in FL, \methodname{} methods are expected to attain certain desirable properties \cite{zeng2021comprehensive,qiu2022applications,ali2021incentive}.
\begin{enumerate}
    \item \textit{Budget Balance (BB)}: The budget balance property should hold, i.e., the total payments for DOs must not surpass the budget allocated by the DCs.
    \item \textit{Collusion Resistant (CR)}: This property imposes that no subgroups of participants can achieve higher profits through collusion or unethical conduct.
    \item \textit{Pareto Efficiency (PE)}: \methodname{} methods must meet the PE requirement when maximizing the social welfare of the entire AFL ecosystem.
    \item \textit{Fairness}: This property means that the entire AFL ecosystem should achieve a predefined fairness notion, such as contribution fairness, regret distribution fairness, or expectation fairness \cite{shi2023fairfed}. 
    \item \textit{Individual Rationality (IR)}: An \methodname{} method is deemed IR only if the profits for all participants are non-negative.
    \item \textit{Incentive Compatibility (IC) / Truthfulness}: Achieving IC/Truthfulness indicates that it is optimal for all participants to truthfully declare their contributions and cost types. Reporting untruthful information does not yield additional gain.
    \item \textit{Computational efficiency (CE)}: This property demands that the incorporated agents must guarantee the completion of the auction process and payment within polynomial time for operational efficiency in AFL. 
\end{enumerate}

\subsection{Experimental Evaluation Metrics}
Experimental evaluation plays a pivotal role in assessing and validating the efficacy of \methodname{} methods. It is instrumental in gauging the performance of these agents under complex settings. The following experimental evaluation metrics are commonly adopted by existing literature to quantitatively measure the effectiveness and impact of \methodname{}:
\begin{enumerate}
    \item \textit{Quality-of-Experience (QoE)}. QoE is expressed as the ratio between FL task completion time to the deadline of the task. It measures the speed at which a DC receives service from a DO, providing insights into the responsiveness and efficiency of the \methodname{} method.
    \item \textit{Utility}. It reflects the utility attained by DCs or DOs during the successful execution of FL tasks. A higher value indicates greater satisfaction with the received results, offering insights into the effectiveness of decisions made by the \methodname{} method. 
    It can be expressed in various forms (e.g., the averaged form or the summation form).
    \item \textit{Task Completion Ratio}. This metric is expressed as the number of successful trades by DCs and is calculated as the ratio of the total number of winning DCs to the total number of DCs in the AFL marketplace. A higher task completion ratio indicates that more FL tasks are successfully allocated to DOs, providing a measure of the efficiency of the \methodname{} method.
    \item \textit{Payment}. Payment for DOs quantifies the financial compensation they received for the successful completion of FL tasks. This metric reflects the economic incentive and compensation provided to DOs, highlighting their contributions to the AFL marketplace under the given \methodname{} method.
    \item \textit{Social welfare}: Social welfare is a comprehensive metric that considers the collective well-being or total utility of all participants in the AFL marketplace, including both DCs and DOs. It provides a holistic measure of the overall effectiveness and fairness of the AFL ecosystem by considering the welfare of all stakeholders.
\end{enumerate}

\section{Promising Future Research Directions}
Through our survey, it can be observed that AFL is still in its early stages of development, with various challenges yet to be addressed. This section delves into potential future directions for this nascent and interdisciplinary field.

\subsection{Dynamic Decision Update}
Existing \methodname{} methods are generally static approaches, represented by linear or non-linear functions. These functions derive their parameters from historical auction data through heuristic techniques. However, these static methods face a challenge when applied to new auctions, as the dynamics of these auctions may differ significantly from historical data. The inherent dynamism of the AFL market poses a considerable obstacle for static bidding methods to achieve desired outcomes in novel auction scenarios consistently.

To address this challenge, incorporating dynamic decision updates for both DOs and DCs, in accordance with the principles of demand-supply economics \cite{nedelec2022learning}, is a promising direction. Such dynamic pricing approaches extend the auctioneer's role as well. A promising avenue for future exploration involves utilizing deep learning approaches to comprehend and model the behaviors of both DOs and DCs. Integrating these learned behaviors into various decision-making processes holds the potential to significantly enhance their utilities, adapting to the evolving dynamics of AFL marketplaces.

\subsection{Multi-Agent Systems}
AFL involves diverse stakeholders, each assuming distinct roles and harboring varied concerns. AFL, at its core, constitutes a multi-agent system (MAS), where intelligent entities interact dynamically within a complex framework. As illustrated in \cite{tang2023multi}, the relationships among DCs add a layer of intricacy, characterized by the simultaneous existence of both competition and cooperation. Moreover, within this ecosystem, the decision-making process of each participant carries direct or indirect repercussions on the choices made by other involved parties. Hence, adopting a MAS perspective to conceptualize AFL to provide a holistic understanding of the intricate interplay among diverse entities is a promising research direction \cite{kraus2023customer}.

\subsection{Preserving Privacy and Improving Security}
Most existing auction-based mechanisms involve third-party entities, such as edge servers acting as auctioneers to manage each auction process. However, relying on third-party entities raises concerns about security and potential privacy breaches \cite{tang2022towards}. To address these challenges, several studies, including \cite{batool2023block,zhang2021incentive,batool2022fl}, utilize blockchain technology to safeguard trading information against tampering by malicious entities. However, implementing an auction algorithm within a blockchain network necessitates sharing private information among stakeholders, potentially giving rise to privacy concerns \cite{tang2022towards}. Moreover, in most existing works, DOs participate in the auction process without directly disclosing their private information, potentially dampening the enthusiasm of DOs. Therefore, a critical challenge arises in ensuring the security and reliability of auction mechanisms, while minimizing the risk of privacy leakage. In addition, it is essential to develop strategies to prevent malicious edge servers from launching attacks on DOs \cite{lyu2020threats}.

\subsection{Online Auction Mechanisms}
The current paradigm of \methodname{}, rooted in traditional auction methods, predominantly operates in an offline mode. This implies that the initiation of auctions relies on having a sufficient number of available bidders. For instance, in \cite{zeng2020fmore}, the model aggregator initiates the process of determining winners once a satisfactory number of bids from DOs is received. In such offline auctions, both the DOs and the DCs may experience prolonged waiting times, even if they do not emerge as the eventual auction winners. This can discourage potential participants from actively engaging in the AFL marketplace. In contrast, online auction \cite{zhang2020online} empowers the auctioneer, DCs and DOs to make real-time decisions, such as selecting winners and determining payments, as soon as a participant joins the auction. Online auctions offer the advantage of overcoming time and space constraints, ultimately resulting in cost savings. Therefore, online auction is a promising research direction for designing stronger incentive mechanisms in AFL.

\subsection{Efficient Contribution Evaluation Methods}
A crucial phase in the auction process involves the selection of the winning DOs, which heavily relies on evaluating the contributions of each DO. The prevailing approach employed by existing \methodname{} methods centers on contribution evaluation methods based on Shapley values. However, as highlighted in \cite{liu2022gtg}, methods grounded in Shapley values are often time-consuming, posing a challenge to the computational efficiency when the system is scaled up. Furthermore, these methods operate under the assumption that DCs and other participants will truthfully assess the contribution of each DO, introducing a potential limitation in scenarios where honesty cannot be guaranteed. Hence, exploring alternative, more efficient contribution evaluation methods is a promising research direction to enhance the efficacy of \methodname{} methods.

\subsection{Explainable AFL}
As indicated by \cite{tang2022towards}, explainability is an important aspect for auctions. Therefore, in the realm of AFL, an intriguing future direction is the advancement of Explainable AFL. This forward-looking approach entails the integration of mechanisms geared towards augmenting the transparency and interpretability of both the auction processes and federated training processes \cite{li2023towards}. The implementation of explainability in AFL holds the potential to foster heightened levels of trust, accountability, comprehensibility and auditability regarding the decision-making processes involved in both the auction and the federated training phases.

\section{Concluding Remarks}
In this paper, we conduct a comprehensive review of \methodname{} methods through a unique multi-tiered taxonomy that organises existing works according to 1) the stakeholders served, 2) the auction mechanism adopted, and 3) the goals of the agents. 
Furthermore, we critically analyze the limitations of current approaches, outline commonly utilized performance evaluation methodologies, and deliberate on promising future directions. 
To the best of our knowledge, it is the first survey on \methodname{}, providing researchers with an accessible guide into this interdisciplinary field.


\bibliographystyle{named}
\bibliography{main}

\begin{thebibliography}{}

\bibitem[\protect\citeauthoryear{Ali and others}{2021}]{ali2021incentive}
Asad Ali et~al.
\newblock Incentive-driven federated learning and associated security challenges: A systematic review, 2021.

\bibitem[\protect\citeauthoryear{Batool \bgroup \em et al.\egroup }{2022}]{batool2022fl}
Zahra Batool, Kaiwen Zhang, and Matthew Toews.
\newblock Fl-mab: client selection and monetization for blockchain-based federated learning.
\newblock In {\em SAC}, pages 299--307, 2022.

\bibitem[\protect\citeauthoryear{Batool \bgroup \em et al.\egroup }{2023}]{batool2023block}
Zahra Batool, Kaiwen Zhang, and Matthew Toews.
\newblock Block-racs: Towards reputation-aware client selection and monetization mechanism for federated learning.
\newblock {\em SAC}, 23(3):49--66, 2023.

\bibitem[\protect\citeauthoryear{Buchbinder and others}{2009}]{buchbinder2009design}
Niv Buchbinder et~al.
\newblock The design of competitive online algorithms via a primal--dual approach.
\newblock {\em FTTCS}, 3(2--3):93--263, 2009.

\bibitem[\protect\citeauthoryear{Cong and others}{2020a}]{cong2020optimal}
Mingshu Cong et~al.
\newblock Optimal procurement auction for cooperative production of virtual products: Vickrey-clarke-groves meet cremer-mclean.
\newblock {\em arXiv preprint arXiv:2007.14780}, 2020.

\bibitem[\protect\citeauthoryear{Cong and others}{2020b}]{cong2020vcg}
Mingshu Cong et~al.
\newblock A vcg-based fair incentive mechanism for federated learning.
\newblock {\em arXiv preprint arXiv:2008.06680}, 2020.

\bibitem[\protect\citeauthoryear{De~Vries and Vohra}{2003}]{de2003combinatorial}
Sven De~Vries and Rakesh~V Vohra.
\newblock Combinatorial auctions: A survey.
\newblock {\em JOC}, 15(3):284--309, 2003.

\bibitem[\protect\citeauthoryear{Deng and others}{2021}]{deng2021fair}
Yongheng Deng et~al.
\newblock Fair: Quality-aware federated learning with precise user incentive and model aggregation.
\newblock In {\em INFOCOM}, 2021.

\bibitem[\protect\citeauthoryear{Fan \bgroup \em et al.\egroup }{2020}]{fan2020hybrid}
Sizheng Fan, Hongbo Zhang, Yuchen Zeng, and Wei Cai.
\newblock Hybrid blockchain-based resource trading system for federated learning in edge computing.
\newblock {\em IOTJ}, 8(4):2252--2264, 2020.

\bibitem[\protect\citeauthoryear{Friedman}{2018}]{friedman2018double}
Daniel Friedman.
\newblock {\em The double auction market: institutions, theories, and evidence}.
\newblock Routledge, 2018.

\bibitem[\protect\citeauthoryear{Gao \bgroup \em et al.\egroup }{2023}]{Gao-et-al:2023ICME}
Yulan Gao, Yansong Zhao, and Han Yu.
\newblock Multi-tier client selection for mobile federated learning networks.
\newblock In {\em ICME}, pages 666--671, 2023.

\bibitem[\protect\citeauthoryear{Gollapudi and others}{2017}]{gollapudi2017profit}
Sreenivas Gollapudi et~al.
\newblock Profit sharing and efficiency in utility games.
\newblock In {\em ESA}, 2017.

\bibitem[\protect\citeauthoryear{Jiao and others}{2020}]{jiao2020toward}
Yutao Jiao et~al.
\newblock Toward an automated auction framework for wireless federated learning services market.
\newblock {\em TMC}, 20(10):3034--3048, 2020.

\bibitem[\protect\citeauthoryear{Kairouz \bgroup \em et al.\egroup }{2021}]{Kairouz-et-al:2021}
Peter Kairouz, H.~Brendan McMahan, et~al.
\newblock Advances and open problems in federated learning.
\newblock {\em FTML}, 14(1-2):1--210, 2021.

\bibitem[\protect\citeauthoryear{Kang and others}{2019}]{kang2019incentive}
Jiawen Kang et~al.
\newblock Incentive mechanism for reliable federated learning: A joint optimization approach to combining reputation and contract theory.
\newblock {\em IOTJ}, 6(6):10700--10714, 2019.

\bibitem[\protect\citeauthoryear{Kang and others}{2020}]{kang2020reliable}
Jiawen Kang et~al.
\newblock Reliable federated learning for mobile networks.
\newblock {\em Wireless Commun}, 27(2):72--80, 2020.

\bibitem[\protect\citeauthoryear{Khan and others}{2020}]{khan2020federated}
Latif~U. Khan et~al.
\newblock Federated learning for edge networks: Resource optimization and incentive mechanism.
\newblock {\em Commun Mag}, 58(10):88--93, 2020.

\bibitem[\protect\citeauthoryear{Kim \bgroup \em et al.\egroup }{2019}]{kim2019blockchained}
Hyesung Kim, Jihong Park, Mehdi Bennis, and Seong-Lyun Kim.
\newblock Blockchained on-device federated learning.
\newblock {\em Commun Lett}, 24(6):1279--1283, 2019.

\bibitem[\protect\citeauthoryear{Kone{\v{c}}n{\`y} \bgroup \em et al.\egroup }{2016}]{konevcny2016federated}
Jakub Kone{\v{c}}n{\`y}, H~Brendan McMahan, Daniel Ramage, and Peter Richt{\'a}rik.
\newblock Federated optimization: Distributed machine learning for on-device intelligence, 2016.

\bibitem[\protect\citeauthoryear{Kraus and others}{2023}]{kraus2023customer}
Sarit Kraus et~al.
\newblock Customer service combining human operators and virtual agents: A call for multidisciplinary ai research.
\newblock 2023.

\bibitem[\protect\citeauthoryear{Le and others}{2020}]{le2020auction}
Tra Huong~Thi Le et~al.
\newblock Auction based incentive design for efficient federated learning in cellular wireless networks.
\newblock In {\em WCNC}, pages 1--6, 2020.

\bibitem[\protect\citeauthoryear{Li \bgroup \em et al.\egroup }{2023}]{li2023towards}
Anran Li, Rui Liu, Ming Hu, Luu~Anh Tuan, and Han Yu.
\newblock Towards interpretable federated learning.
\newblock {\em arXiv preprint arXiv:2302.13473}, 2023.

\bibitem[\protect\citeauthoryear{Lim and others}{2020}]{lim2020incentive}
Wei Yang~Bryan Lim et~al.
\newblock Incentive mechanism design for resource sharing in collaborative edge learning.
\newblock {\em arXiv preprint arXiv:2006.00511}, 2020.

\bibitem[\protect\citeauthoryear{Liu and others}{2020}]{Liu-et-al:2020FedVision}
Yang Liu et~al.
\newblock {FedVision}: An online visual object detection platform powered by federated learning.
\newblock In {\em IAAI}, pages 13172--13179, 2020.

\bibitem[\protect\citeauthoryear{Liu and others}{2022a}]{Liu-et-al:2022IAAI}
Zelei Liu et~al.
\newblock Contribution-aware federated learning for smart healthcare.
\newblock In {\em IAAI}, pages 12396--12404, 2022.

\bibitem[\protect\citeauthoryear{Liu and others}{2022b}]{liu2022gtg}
Zelei Liu et~al.
\newblock {GTG-Shapley}: Efficient and accurate participant contribution evaluation in federated learning.
\newblock {\em TIST}, 13(4):1--21, 2022.

\bibitem[\protect\citeauthoryear{Lu \bgroup \em et al.\egroup }{2023}]{lu2023auction}
Renhao Lu, Weizhe Zhang, Yan Wang, Qiong Li, Xiaoxiong Zhong, Hongwei Yang, and Desheng Wang.
\newblock Auction-based cluster federated learning in mobile edge computing systems.
\newblock {\em TPDS}, 34(4):1145--1158, 2023.

\bibitem[\protect\citeauthoryear{Lyu \bgroup \em et al.\egroup }{2020}]{lyu2020threats}
Lingjuan Lyu, Han Yu, and Qiang Yang.
\newblock Threats to federated learning: A survey.
\newblock {\em arXiv preprint arXiv:2003.02133}, 2020.

\bibitem[\protect\citeauthoryear{Mai \bgroup \em et al.\egroup }{2022}]{mai2022automatic}
Tianle Mai, Haipeng Yao, Jiaqi Xu, Ni~Zhang, Qifeng Liu, and Song Guo.
\newblock Automatic double-auction mechanism for federated learning service market in internet of things.
\newblock {\em TNSE}, 9(5):3123--3135, 2022.

\bibitem[\protect\citeauthoryear{McMahan and others}{2017}]{mcmahan2017communication}
Brendan McMahan et~al.
\newblock Communication-efficient learning of deep networks from decentralized data.
\newblock In {\em AISTATS}, pages 1273--1282, 2017.

\bibitem[\protect\citeauthoryear{Mnih and others}{2015}]{mnih2015human}
Volodymyr Mnih et~al.
\newblock Human-level control through deep reinforcement learning.
\newblock {\em Nature}, 518(7540):529--533, 2015.

\bibitem[\protect\citeauthoryear{Myerson}{1998}]{myerson1998population}
Roger~B Myerson.
\newblock Population uncertainty and poisson games.
\newblock {\em IJGT}, 27:375--392, 1998.

\bibitem[\protect\citeauthoryear{Nedelec and others}{2022}]{nedelec2022learning}
Thomas Nedelec et~al.
\newblock Learning in repeated auctions.
\newblock {\em FTML}, 15(3):176--334, 2022.

\bibitem[\protect\citeauthoryear{Ng and others}{2020a}]{ng2020communication}
Jer~Shyuan Ng et~al.
\newblock Communication-efficient federated learning in {UAV}-enabled iov: a joint auction-coalition approach.
\newblock In {\em GLOBECOM}, pages 1--6, 2020.

\bibitem[\protect\citeauthoryear{Ng and others}{2020b}]{ng2020joint}
Jer~Shyuan Ng et~al.
\newblock Joint auction-coalition formation framework for communication-efficient federated learning in {UAV}-enabled internet of vehicles.
\newblock {\em TIST}, 22(4):2326--2344, 2020.

\bibitem[\protect\citeauthoryear{Parsons and others}{2011}]{parsons2011auctions}
Simon Parsons et~al.
\newblock Auctions and bidding: A guide for computer scientists.
\newblock {\em CSUR}, 43(2):1--59, 2011.

\bibitem[\protect\citeauthoryear{Qiu and others}{2022}]{qiu2022applications}
Houming Qiu et~al.
\newblock Applications of auction and mechanism design in edge computing: A survey.
\newblock {\em TCCN}, 8(2):1034--1058, 2022.

\bibitem[\protect\citeauthoryear{Roy and others}{2021}]{roy2021distributed}
Palash Roy et~al.
\newblock Distributed task allocation in mobile device cloud exploiting federated learning and subjective logic.
\newblock {\em JSA}, 113(2), 2021.

\bibitem[\protect\citeauthoryear{Seo and others}{2021}]{seo2021auction}
Eunil Seo et~al.
\newblock Auction-based federated learning using software-defined networking for resource efficiency.
\newblock In {\em CNSM}, pages 42--48, 2021.

\bibitem[\protect\citeauthoryear{Seo and others}{2022}]{seo2022resource}
Eunil Seo et~al.
\newblock Resource-efficient federated learning with non-iid data: An auction theoretic approach.
\newblock {\em IOTJ}, 9(24):25506--25524, 2022.

\bibitem[\protect\citeauthoryear{Shi and Yu}{2023}]{shi2023fairfed}
Yuxin Shi and Han Yu.
\newblock Fairness-aware client selection for federated learning.
\newblock In {\em ICME}, 2023.

\bibitem[\protect\citeauthoryear{Tan and Yu}{2023}]{tan2023hire}
Xavier Tan and Han Yu.
\newblock Hire when you need to: Gradual participant recruitment for auction-based federated learning.
\newblock {\em arXiv preprint arXiv:2310.02651}, 2023.

\bibitem[\protect\citeauthoryear{Tan \bgroup \em et al.\egroup }{2023}]{tan2023reputation}
Xavier Tan, Wei Yang~Bryan Lim, Dusit Niyato, and Han Yu.
\newblock Reputation-aware opportunistic budget optimization for auction-based federation learning.
\newblock In {\em IJCNN}, pages 1--8, 2023.

\bibitem[\protect\citeauthoryear{Tang and Yu}{2022}]{tang2022towards}
Xiaoli Tang and Han Yu.
\newblock Towards trustworthy ai-empowered real-time bidding for online advertisement auctioning.
\newblock {\em arXiv preprint arXiv:2210.07770}, 2022.

\bibitem[\protect\citeauthoryear{Tang and Yu}{2023a}]{tang2023competitive}
Xiaoli Tang and Han Yu.
\newblock Competitive-cooperative multi-agent reinforcement learning for auction-based federated learning.
\newblock In {\em IJCAI}, 2023.

\bibitem[\protect\citeauthoryear{Tang and Yu}{2023b}]{tang2023multi}
Xiaoli Tang and Han Yu.
\newblock Multi-session budget optimization for forward auction-based federated learning.
\newblock {\em arXiv preprint arXiv:2311.12548}, 2023.

\bibitem[\protect\citeauthoryear{Tang and Yu}{2023c}]{tang2023utility}
Xiaoli Tang and Han Yu.
\newblock Utility-maximizing bidding strategy for data consumers in auction-based federated learning.
\newblock In {\em ICME}, 2023.

\bibitem[\protect\citeauthoryear{Thi~Le and others}{2021}]{le2021incentive}
Tra~Huong Thi~Le et~al.
\newblock An incentive mechanism for federated learning in wireless cellular networks: An auction approach.
\newblock {\em TWC}, 20(8):4874--4887, 2021.

\bibitem[\protect\citeauthoryear{Tu and others}{2022}]{tu2022incentive}
Xuezhen Tu et~al.
\newblock Incentive mechanisms for federated learning: From economic and game theoretic perspective.
\newblock {\em TCCN}, 8(3):1566--1593, 2022.

\bibitem[\protect\citeauthoryear{Vickrey}{1961}]{vickrey1961counterspeculation}
William Vickrey.
\newblock Counterspeculation, auctions, and competitive sealed tenders.
\newblock {\em JF}, 16(1):8--37, 1961.

\bibitem[\protect\citeauthoryear{Wang and others}{2023}]{wang2023toward}
Fei Wang et~al.
\newblock Toward sustainable ai: Federated learning demand response in cloud-edge systems via auctions.
\newblock In {\em INFOCOM}, pages 1--10, 2023.

\bibitem[\protect\citeauthoryear{Wu and others}{2023}]{wu2023long}
Leijie Wu et~al.
\newblock Long-term adaptive vcg auction mechanism for sustainable federated learning with periodical client shifting.
\newblock {\em TMC}, 2023.

\bibitem[\protect\citeauthoryear{Xu and others}{2021}]{xu2021bandwidth}
Jie Xu et~al.
\newblock Bandwidth allocation for multiple federated learning services in wireless edge networks.
\newblock {\em TWC}, 21(4):2534--2546, 2021.

\bibitem[\protect\citeauthoryear{Xu and others}{2023}]{xu2023cab}
Bo~Xu et~al.
\newblock Cab: a combinatorial-auction-and-bargaining-based federated learning incentive mechanism.
\newblock {\em WWW}, pages 1--22, 2023.

\bibitem[\protect\citeauthoryear{Yang and others}{2023}]{yang2023bara}
Yunchao Yang et~al.
\newblock Bara: Efficient incentive mechanism with online reward budget allocation in cross-silo federated learning.
\newblock {\em arXiv preprint arXiv:2305.05221}, 2023.

\bibitem[\protect\citeauthoryear{Yang \bgroup \em et al.\egroup }{2019}]{yang2019federated}
Qiang Yang, Yang Liu, Tianjian Chen, and Yongxin Tong.
\newblock Federated machine learning: Concept and applications.
\newblock {\em TIST}, 10(2):12:1--12:19, 2019.

\bibitem[\protect\citeauthoryear{Yuan and others}{2021}]{yuan2021incentivizing}
Yulan Yuan et~al.
\newblock Incentivizing federated learning under long-term energy constraint via online randomized auctions.
\newblock {\em TWC}, 21(7):5129--5144, 2021.

\bibitem[\protect\citeauthoryear{Zeng and others}{2021}]{zeng2021comprehensive}
Rongfei Zeng et~al.
\newblock A comprehensive survey of incentive mechanism for federated learning.
\newblock {\em arXiv preprint arXiv:2106.15406}, 2021.

\bibitem[\protect\citeauthoryear{Zeng \bgroup \em et al.\egroup }{2020}]{zeng2020fmore}
Rongfei Zeng, Shixun Zhang, Jiaqi Wang, and Xiaowen Chu.
\newblock Fmore: An incentive scheme of multi-dimensional auction for federated learning in {MEC}.
\newblock In {\em ICDCS}, pages 278--288, 2020.

\bibitem[\protect\citeauthoryear{Zhang and others}{2020}]{zhang2020online}
Jixian Zhang et~al.
\newblock An online auction mechanism for time-varying multidimensional resource allocation in clouds.
\newblock {\em FGCS}, 111:27--38, 2020.

\bibitem[\protect\citeauthoryear{Zhang \bgroup \em et al.\egroup }{2021}]{zhang2021incentive}
Jingwen Zhang, Yuezhou Wu, and Rong Pan.
\newblock Incentive mechanism for horizontal federated learning based on reputation and reverse auction.
\newblock In {\em WWW}, page 947–956, 2021.

\bibitem[\protect\citeauthoryear{Zhang \bgroup \em et al.\egroup }{2022a}]{zhang2022auction}
Jingwen Zhang, Yuezhou Wu, and Rong Pan.
\newblock Auction-based ex-post-payment incentive mechanism design for horizontal federated learning with reputation and contribution measurement.
\newblock {\em arXiv preprint arXiv:2201.02410}, 2022.

\bibitem[\protect\citeauthoryear{Zhang \bgroup \em et al.\egroup }{2022b}]{zhang2022online}
Jingwen Zhang, Yuezhou Wu, and Rong Pan.
\newblock Online auction-based incentive mechanism design for horizontal federated learning with budget constraint.
\newblock {\em arXiv preprint}, page 2201.09047, 2022.

\bibitem[\protect\citeauthoryear{Zhao and others}{2018}]{zhao2018federated}
Yue Zhao et~al.
\newblock Federated learning with non-iid data.
\newblock {\em arXiv preprint arXiv:1806.00582}, 2018.

\bibitem[\protect\citeauthoryear{Zhou \bgroup \em et al.\egroup }{2021}]{zhou2021truthful}
Ruiting Zhou, Jinlong Pang, Zhibo Wang, John~CS Lui, and Zongpeng Li.
\newblock A truthful procurement auction for incentivizing heterogeneous clients in federated learning.
\newblock In {\em ICDCS}, pages 183--193, 2021.

\end{thebibliography}

\end{document}